\documentclass[runningheads]{llncs}
\usepackage{comment}
\usepackage{graphicx}
\usepackage{epsfig}
\usepackage{amsmath}
\usepackage{amssymb}
\usepackage{multirow}
\usepackage{cite}
\usepackage{xcolor}
\usepackage{subcaption}
\usepackage{url}

\begin{document}
\title{Joint Modeling of Chest Radiographs and Radiology Reports for Pulmonary Edema Assessment}
\titlerunning{Joint Modeling of Chest Radiographs and Radiology Reports}
%
\author{Geeticka Chauhan$^{1}$\thanks{Co-first authors}, Ruizhi Liao$^{1}$\textsuperscript{\thefootnote}, William Wells$^{1,2}$, Jacob Andreas$^{1}$, \\
Xin Wang$^{3}$, Seth Berkowitz$^{4}$, Steven Horng$^{4}$, \\
Peter Szolovits$^{1}$, and Polina Golland$^{1}$}
%
\authorrunning{G. Chauhan, R. Liao et al.}
%
\institute{$^{1}$  Massachusetts Institute of Technology, Cambridge, MA, USA\\
$^{2}$ Brigham and Women's Hospital, Harvard Medical School, Boston, MA, USA\\
$^{3}$ Philips Research North America, Cambridge, MA, USA\\
$^{4}$ Beth Israel Deaconess Medical Center, Harvard Medical School, Boston, MA, USA}
\maketitle              
\begin{abstract}
We propose and demonstrate a novel machine learning algorithm that assesses pulmonary edema severity from chest radiographs. While large publicly available datasets of chest radiographs and free-text radiology reports exist, only limited numerical edema severity labels can be extracted from radiology reports. This is a significant challenge in learning such models for image classification. To take advantage of the rich information present in the radiology reports, we develop a neural network model that is trained on both images and free-text to assess pulmonary edema severity from chest radiographs at inference time. Our experimental results suggest that the joint image-text representation learning improves the performance of pulmonary edema assessment compared to a supervised model trained on images only. We also show the use of the text for explaining the image classification by the joint model. To the best of our knowledge, our approach is the first to leverage free-text radiology reports for improving the image model performance in this application. Our code is available at: \url{https://github.com/RayRuizhiLiao/joint_chestxray}.


\end{abstract}
\section{Introduction}

We present a novel approach to training machine learning models for assessing pulmonary edema severity from chest radiographs by jointly learning representations from the images (chest radiographs) and their associated radiology reports. Pulmonary edema is the most common reason patients with acute congestive heart failure~(CHF) seek care in hospitals~\cite{gheorghiade2010assessing, hunt20092009, adams2005characteristics}. The treatment success in acute CHF cases depends crucially on effective management of patient fluid status, which in turn requires pulmonary edema quantification, rather than detecting its mere absence or presence. 

Chest radiographs are commonly acquired to assess pulmonary edema in routine clinical practice. Radiology reports capture radiologists' impressions of the edema severity in the form of unstructured text. While the chest radiographs possess ground-truth information about the disease, they are often time intensive (and therefore expensive) for manual labeling. Therefore, labels extracted from reports are used as a proxy for ground-truth image labels. Only limited numerical edema severity labels can be extracted from the reports, which limits the amount of labeled image data we can learn from. This presents a significant challenge for learning accurate image-based models for edema assessment. To improve the performance of the image-based model and allow leveraging larger amount of training data, we make use of free-text reports to include rich information about radiographic findings and reasoning of pathology assessment. We incorporate free-text information associated with the images by including them during our training process. 

We propose a neural network model that jointly learns from images and free-text to quantify pulmonary edema severity from images (chest radiographs). At training time, the model learns from a large number of chest radiographs and their associated radiology reports, with a limited number of numerical edema severity labels. At inference time, the model computes edema severity given the input image. While the model can also make predictions from reports, our main interest is to leverage free-text information during training to improve the accuracy of image-based inference. Compared to prior work in the image-text domain that fuses image and text features~\cite{ben2017mutan}, our goal is to decouple the two modalities during inference to construct an accurate image-based model.



Prior work in assessing pulmonary edema severity from chest radiographs has focused on using image data only~\cite{liao2019semi, horng2020deep}. To the best of our knowledge, ours is the first method to leverage the free-text radiology reports for improving the image model performance in this application. Our experimental results demonstrate that the joint representation learning framework improves the accuracy of edema severity estimates over a purely image-based model on a fully labeled subset of the data (supervised). The joint learning framework uses a ranking-based criterion~\cite{harwath2016unsupervised, chechik2010large}, allowing for training the model on a larger dataset of unlabeled images and reports. This semi-supervised modification demonstrates a further improvement in accuracy. Additional advantages of our joint learning framework are 1) allowing for the image and text models to be decoupled at inference time, and 2) providing textual explanations for image classification in the form of saliency highlights in the radiology reports.


\paragraph{\textbf{Related Work.}}
The ability of neural networks to learn effective feature representations from images and text has catalyzed the recent surge of interest in joint image-text modeling. In supervised learning, tasks such as image captioning have leveraged a recurrent visual attention mechanism using recurrent neural networks~(RNNs) to improve captioning performance~\cite{xu2015show}. The TieNet used this attention-based text embedding framework for pathology detection from chest radiographs~\cite{wang2018tienet}, which was further improved by introducing a global topic vector and transfer learning~\cite{xue2019improved}. A similar image-text embedding setup has been employed for chest radiograph (image) annotations~\cite{moradi2018bimodal}. In unsupervised learning, training a joint global embedding space for visual object discovery has recently been shown to capture relevant structure~\cite{harwath2018jointly}. All of these models used RNNs for encoding text features. More recently, transformers such as the BERT model~\cite{devlin2018bert} have shown the ability to capture richer contextualized word representations using self-attention and have advanced the state-of-the-art in nearly every language processing task compared to variants of RNNs. Our setup, while similar to \cite{wang2018tienet} and \cite{harwath2018jointly}, uses a series of residual blocks~\cite{he2016deep} to encode the image representation and uses the BERT model to encode the text representation. We use the radiology reports during training only, to improve the image-based model's performance. This is in contrast to visual question answering~\cite{antol2015vqa, lu2016hierarchical, anderson2018bottom}, where inference is performed on an image-text pair, and image/video captioning~\cite{xu2015show, plummer2017enhancing, vasudevan2017query,jing2017automatic}, where the model generates text from the input image.

\section{Data}
\label{sec:data}
For training and evaluating our model, we use the MIMIC-CXR dataset v2.0~\cite{johnson2019mimic}, consisting of 377,110 chest radiographs associated with 227,835 radiology reports. The data was collected in routine clinical practice, and each report is associated with one or more images. We limited our study to 247,425 frontal-view radiographs. 


\paragraph{\textbf{Regex Labeling.}} We extracted pulmonary edema severity labels from the associated radiology reports using regular expressions (regex) with negation detection~\cite{chapman2001simple}. The keywords of each severity level (“none”=0, “vascular congestion”=1, “interstitial edema”=2, and “alveolar edema”=3) are summarized in the supplementary materials. In order to limit confounding keywords from other disease processes, we limited the label extraction to patients with congestive heart failure (CHF) based on their ED ICD-9 diagnosis code in the MIMIC dataset~\cite{goldberger2000physiobank}. Cohort selection by diagnosis code for CHF was previously validated by manual chart review. This resulted in 16,108 radiology reports. Regex labeling yielded 6,710 labeled reports associated with 6,743 frontal-view images\footnote{The numbers of images of the four severity levels are 2883, 1511, 1709, and 640 respectively.}. Hence, our dataset includes 247,425 image-text pairs, 6,743 of which are of CHF patients with edema severity labels. Note that some reports are associated with more than one image, so one report may appear in more than one image-text pair. 

\begin{figure*}[!t]
	\centering
	\includegraphics[width=1\linewidth]{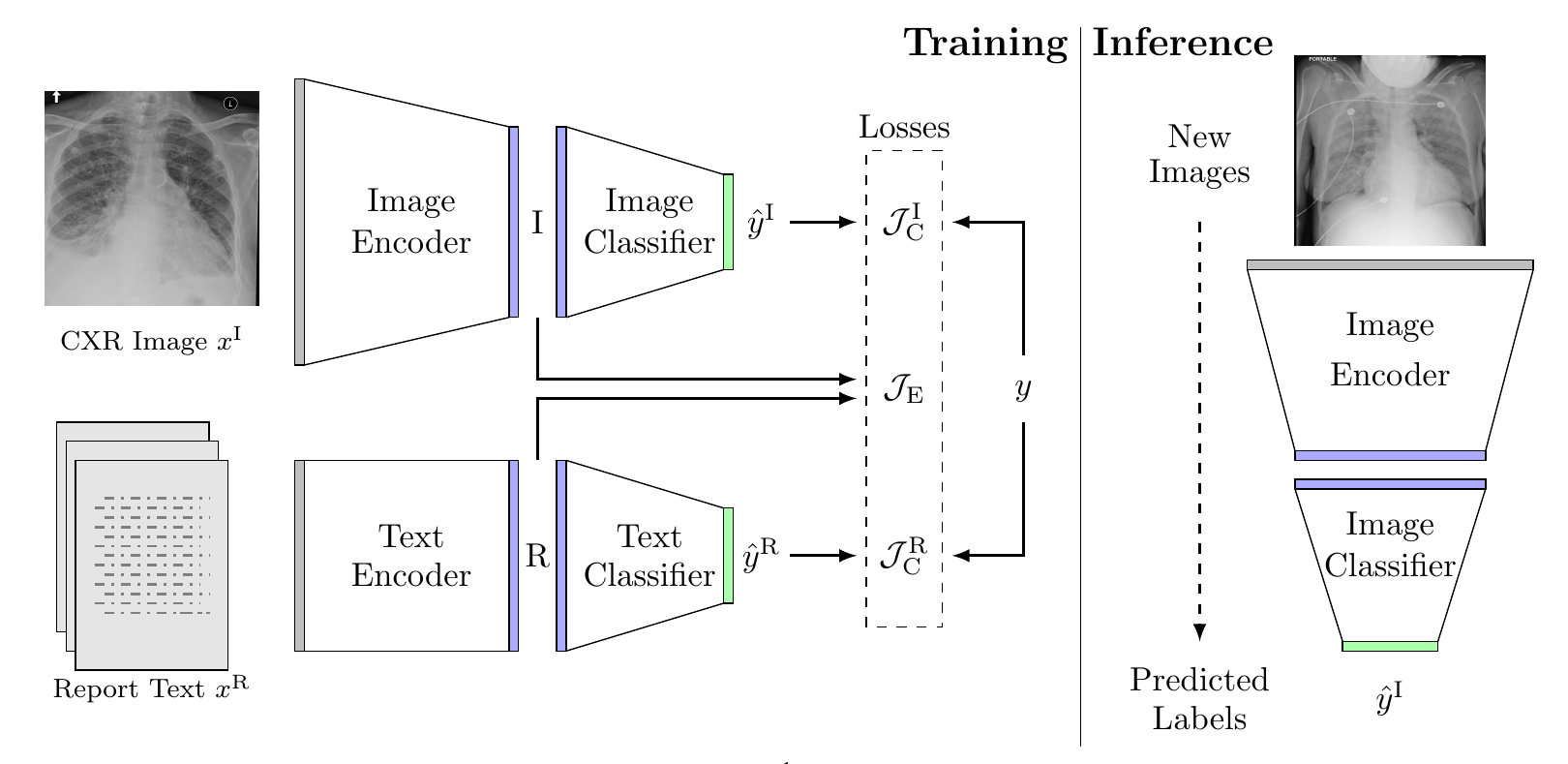}
	\caption{The architecture of our joint model, along with an example chest radiograph~$x^{\text{I}}$ and its associated radiology report~$x^{\text{R}}$. At training time, the model predicts the edema severity level from images and text through their respective encoders and classifiers, and compares the predictions with the labels. The joint embedding loss~$\mathcal{J_\text{E}}$ associates image embeddings~$\text{I}$ with text embeddings~$\text{R}$ in the joint embedding space. At inference time, the image stream and the text stream are decoupled and only the image stream is used. Given a new chest radiograph (image), the image encoder and classifier compute its edema severity level.}
	\label{fig:joint-model}
\end{figure*}

\section{Methods}
\label{sec:methods}

Let $x^{\text{I}}$ be a 2D chest radiograph, $x^{\text{R}}$ be the free-text in a radiology report, and $y\in{\{0,1,2,3\}}$ be the corresponding edema severity label. Our dataset includes a set of $N$ image-text pairs $\text{X}=\{\text{x}_j\}_{j=1}^{N}$, where $\text{x}_j=(\text{x}^{\text{I}}_{j}, \text{x}^{\text{R}}_{j})$. The first $N_{\text{L}}$ image-text pairs are annotated with severity labels $\text{Y}=\{\text{y}_{j}\}_{j=1}^{N_\text{L}}$. Here we train a joint model that constructs an image-text embedding space, where an image encoder and a text encoder are used to extract image features and text features separately (Fig.~\ref{fig:joint-model}). Two classifiers are trained to classify the severity labels independently from the image features and from the text features. This setup enables us to decouple the image classification and the text classification at inference time. Learning the two representations jointly at training time improves the performance of the image model. 

\paragraph{\textbf{Joint Representation Learning.}} We apply a ranking-based criterion~\cite{chechik2010large, harwath2016unsupervised} for training the image encoder and the text encoder parameterized by~$\theta_\text{E}^\text{I}$ and~$\theta_\text{E}^\text{R}$ respectively, to learn image and text feature representations~$I(x^{\text{I}};\theta_\text{E}^\text{I})$ and~$R(x^{\text{R}};\theta_\text{E}^\text{R})$. Specifically, given an image-text pair~$(\text{x}^{\text{I}}_{j}, \text{x}^{\text{R}}_{j})$, we randomly select an impostor image~$\text{x}^{\text{I}}_{s(j)}$ and an impostor  report~$\text{x}^{\text{R}}_{s(j)}$ from~$\text{X}$. This selection is generated at the beginning of each training epoch. Map $s(j)$ produces a random permutation of $\{1,2,...,N\}$.


We encourage the feature representations between a matched pair~$(\text{I}_{j}, \text{R}_{j})$ to be ``closer" than those between mismatched pairs~$(\text{I}_{s(j)}, \text{R}_{j})$ and~$(\text{I}_{j}, \text{R}_{s(j)})$ in the joint embedding space. Direct minimization of the distance between $I$ and $R$ could end up pushing the image and text features into a small cluster in the embedding space. Instead, we encourage matched image-text features to be close while spreading out all feature representations in the embedding space for downstream classification by constructing an appropriate loss function: 
\begin{align}
\mathcal{J_\text{E}}(\theta_\text{E}^\text{I}, \theta_\text{E}^\text{R}; \text{x}_j, \text{x}_{s(j)})
= & \text{max}(0, \text{Sim}(\text{I}_{j}, \text{R}_{s(j)})-\text{Sim}(\text{I}_{j}, \text{R}_{j})+\eta) \nonumber \\
& + \text{max} (0, \text{Sim}(\text{I}_{s(j)}, \text{R}_{j})-\text{Sim}(\text{I}_{j}, \text{R}_{j})+\eta),
\label{eq:joint_loss}
\end{align}
where $\text{Sim}(\cdot, \cdot)$ is the similarity measurement of two feature representations in the joint embedding space and $\eta$ is a margin parameter that is set to~$|\text{y}_j-\text{y}_{s(j)}|$ when both $j\leqslant N_\text{L}$ and $s(j)\leqslant N_\text{L}$; otherwise, $\eta=0.5$. The margin is determined by the difference due to the mismatch, if both labels are known; otherwise the margin is a constant.

\paragraph{\textbf{Classification.}} We employ two fully connected layers (with the same neural network architecture) on the joint embedding space to assess edema severity from the image and the report respectively. For simplicity, we treat the problem as multi-class classification, i.e. the classifiers' outputs $\hat{y}^\text{I}(\text{I}; \theta_\text{C}^\text{I})$ and $\hat{y}^\text{R}(\text{R}; \theta_\text{C}^\text{R})$ are encoded as one-hot 4-dimensional vectors. We use cross entropy as the loss function for training the classifiers and the encoders on the labeled data:
\begin{align}
\mathcal{J_\text{C}}(\theta_\text{E}^\text{I}, \theta_\text{E}^\text{R}, \theta_\text{C}^\text{I}, \theta_\text{C}^\text{R}; \text{x}_{j}, \text{y}_{j}) 
= & - \sum_{i=0}^{3} {\text{y}_{ji}}\log \hat{y_i}^\text{I}(\text{I}_{j}(x^{\text{I}}_{j};\theta_\text{E}^\text{I}); \theta_\text{C}^\text{I}) \nonumber \\
& - \sum_{i=0}^{3} {\text{y}_{ji}}\log \hat{y_i}^\text{R}(\text{R}_{j}(x^{\text{R}}_{j};\theta_\text{E}^\text{R}); \theta_\text{C}^\text{R}),
\label{eq:classification_loss}
\end{align}
i.e., minimizing the cross entropy also affects the encoder parameters.  

\paragraph{\textbf{Loss Function.}} Combining Eq.~(\ref{eq:joint_loss}) and Eq.~(\ref{eq:classification_loss}), we obtain the loss function for training the joint model:
\begin{align}
\mathcal{J}(\theta_\text{E}^\text{I}, \theta_\text{E}^\text{R}, \theta_\text{C}^\text{I}, \theta_\text{C}^\text{R}; \text{X}, \text{Y}) 
= & \sum_{j=1}^{N} \mathcal{J_\text{E}}(\theta_\text{E}^\text{I}, \theta_\text{E}^\text{R}; \text{x}_j, \text{x}_{s(j)}) + \sum_{j=1}^{N_\text{L}} \mathcal{J_\text{C}}(\theta_\text{E}^\text{I}, \theta_\text{E}^\text{R}, \theta_\text{C}^\text{I}, \theta_\text{C}^\text{R}; \text{x}_{j}, \text{y}_{j}).
\label{eq:loss_function}
\end{align}

\paragraph{\textbf{Implementation Details.}}
The image encoder is implemented as a series of residual blocks~\cite{he2016deep}, the text encoder is a BERT model that uses the beginner \texttt{[CLS]} token's hidden unit size of 768 and maximum sequence length of 320~\cite{devlin2018bert}. The image encoder is trained from a random initialization, while the BERT model is fine-tuned during the training of the joint model. The BERT model parameters are initialized using pre-trained weights on scientific text~\cite{beltagy2019scibert}. The image features and the text features are represented as 768-dimensional vectors in the joint embedding space. The two classifiers are both 768-to-4 fully connected layers. The neural network architecture is provided in the supplementary materials.

We employ the stochastic gradient-based optimization procedure AdamW~\cite{wolf2019huggingface} to minimize the loss in Eq.~(\ref{eq:loss_function}) and use a warm-up linear scheduler~\cite{attention2017allyouneed} for the learning rate. The model is trained on all the image-text pairs by optimizing the first term in Eq.~(\ref{eq:loss_function}) for 10 epochs and then trained on the labeled image-text pairs by optimizing Eq.~(\ref{eq:loss_function}) for 50 epochs. The mini-batch size is 4. We use dot product as the similarity metric in Eq.~(\ref{eq:joint_loss}). The dataset is split into training and test sets. All the hyper-parameters are selected based on the results from 5-fold cross validation within the training set.

\section{Experiments} 
\label{sec:experiments}
\paragraph{\textbf{Data Preprocessing.}} The size of the chest radiographs varies and is around 3000$\times$3000 pixels. We randomly translate and rotate the images on the fly during training and crop them to 2048$\times$2048 pixels as part of data augmentation. We maintain the original image resolution to capture the subtle differences in the images between different levels of pulmonary edema severity. For the radiology reports, we extract the \textit{impressions}, \textit{findings}, \textit{conclusion} and \textit{recommendation} sections. If none of these sections are present in the report, we use the \textit{final report} section. We perform tokenization of the text using ScispaCy \cite{scispacy2019} before providing it to the BERT tokenizer. 

\paragraph{\textbf{Expert Labeling.}} For evaluating our model, we randomly selected 531 labeled image-text pairs (corresponding to 485 reports) for expert annotation. A board-certified radiologist and two domain experts reviewed and corrected the regex labels of the reports. We use the expert labels for model testing. The overall accuracy of the regex labels (positive predictive value compared against the expert labels) is 89\%.  The other 6,212 labeled image-text pairs and around 240K unlabeled image-text pairs were used for training. There is no patient overlap between the training set and the test set. 

\paragraph{\textbf{Model Evaluation.}} 
We evaluated variants of our model and training regimes as follows: 

\begin{itemize}
    \item \textbf{image-only}: An image-only model with the same architecture as the image stream in our joint model. We trained the image model in isolation on the 6,212 labeled images.
    \item A joint image-text model trained on the 6,212 labeled image-text pairs only. We compare two alternatives to the joint representation learning loss:
    \begin{itemize}
        \item \textbf{ranking-dot}, \textbf{ranking-l2}, \textbf{ranking-cosine}: the ranking based criterion in Eq.~(\ref{eq:joint_loss}) with $\text{Sim}(I, R)$ defined as one of the dot product $I^\top R$, the reciprocal of euclidean distance $-\| I - R \|$, and the cosine similarity $\frac{I^\top R}{ \|I\| . \|R\|}$;
        \item \textbf{dot}, \textbf{l2}, \textbf{cosine}: direct minimization on the similarity metrics without the ranking based criterion.
    \end{itemize}
    \item \textbf{ranking-dot-semi}: A joint image-text model trained on the 6,212 labeled and the 240K unlabeled image-text pairs in a semi-supervised fashion, using the ranking based criterion with dot product in Eq.~(\ref{eq:joint_loss}). Dot product is selected for the ranking-based loss based on cross-validation experiments on the supervised data comparing ranking-dot, ranking-l2, ranking-cosine, dot, l2, and cosine. 
\end{itemize}

All reported results are compared against the expert labels in the test set. The image portion of the joint model is decoupled for testing, and the reported results are predicted from images only. To optimize the baseline performance, we performed a separate hyper-parameter search for the \texttt{image-only} model using 5-fold cross validation (while holding out the test set). 

We use the area under the ROC (AUC) and macro-averaged F1-scores (macro-F1) for our model evaluation. We dichotomize the severity levels and report 3 comparisons (0 \textit{vs} 1,2,3; 0,1 \textit{vs} 2,3; and 0,1,2 \textit{vs} 3), since these 4 classes are ordinal (e.g., $\mathbb{P}(\text{severity}=0 ~\text{or}~ 1)=\hat{y}^\text{I}_0+\hat{y}^\text{I}_1$, $\mathbb{P}(\text{severity}=2~\text{or}~3)=\hat{y}^\text{I}_2+\hat{y}^\text{I}_3$).




\paragraph{\textbf{Results.}}
Table~\ref{table:initial-exp} reports the performance statistics for all similarity measures. The findings are consistent with our cross-validation results: the ranking based criterion offers significant improvement when it is combined with the dot product as the similarity metric.

Table~\ref{table:final-exp} reports the performance of the optimized baseline model (\texttt{image-only}) and two variants of the joint model (\texttt{ranking-dot} and \texttt{ranking-dot-semi}). We observe that when the joint model learns from the large number of unlabeled image-text pairs, it achieves the best performance. The unsupervised learning minimizes the ranking-based loss in Eq.~(\ref{eq:joint_loss}), which does not depend on availability of labels. 

It is not surprising that the model is better at differentiating the severity level 3 than other severity categories, because level 3 has the most distinctive radiographic features in the images. 

\begin{table}[!htb]
  \centering
  \begin{tabular}{|l|r|r|r|r|}
  \hline
  	Method                 & AUC (0 \textit{vs} 1,2,3)     & AUC (0,1 \textit{vs} 2,3)     & AUC (0,1,2 \textit{vs} 3)     &
  	  macro-F1 \\ \hline
    \texttt{l2}             & 0.78              & 0.76              & 0.83  &
    0.42 \\
    \texttt{ranking-l2}     & 0.77              & 0.75              & 0.80  &
    0.43 \\
    \texttt{cosine}         & 0.77              & 0.75              & 0.81  &
    0.44 \\
    \texttt{ranking-cosine} & 0.77              & 0.72              & 0.83 &
    0.41 \\ 
    \texttt{dot}            & 0.65              & 0.63              & 0.61    &
    0.15 \\
    \textbf{ranking-dot}    & \textbf{0.80}      & \textbf{0.78}     & \textbf{0.87}  &
    \textbf{0.45} \\ \hline
  \end{tabular}
  \vspace{.1in}
  \caption{Performance statistics for all similarity measures.}
  \label{table:initial-exp}

  \vspace{.1in}

  \begin{tabular}{|l|r|r|r|r|}
  \hline
  	Method                 & AUC (0 \textit{vs} 1,2,3)     & AUC (0,1 \textit{vs} 2,3)     & AUC (0,1,2 \textit{vs} 3)     &
  	  macro-F1 \\ \hline
  	\texttt{image-only}         & 0.74                 & 0.73                 & 0.78   &
  	  0.43   \\
    \texttt{ranking-dot}        & 0.80               & 0.78              & 0.87  &
  	  0.45   \\
    \textbf{ranking-dot-semi}   & \textbf{0.82}     & \textbf{0.81}     & \textbf{0.90}   &
  	 \textbf{0.51}   \\ \hline
  \end{tabular}
  \vspace{.1in}
  \caption{Performance statistics for the two variants of our joint model and the baseline image model.}
  \label{table:final-exp}
\end{table}

\begin{figure*}[!hb]
	\centering
	\includegraphics[width=1\linewidth]{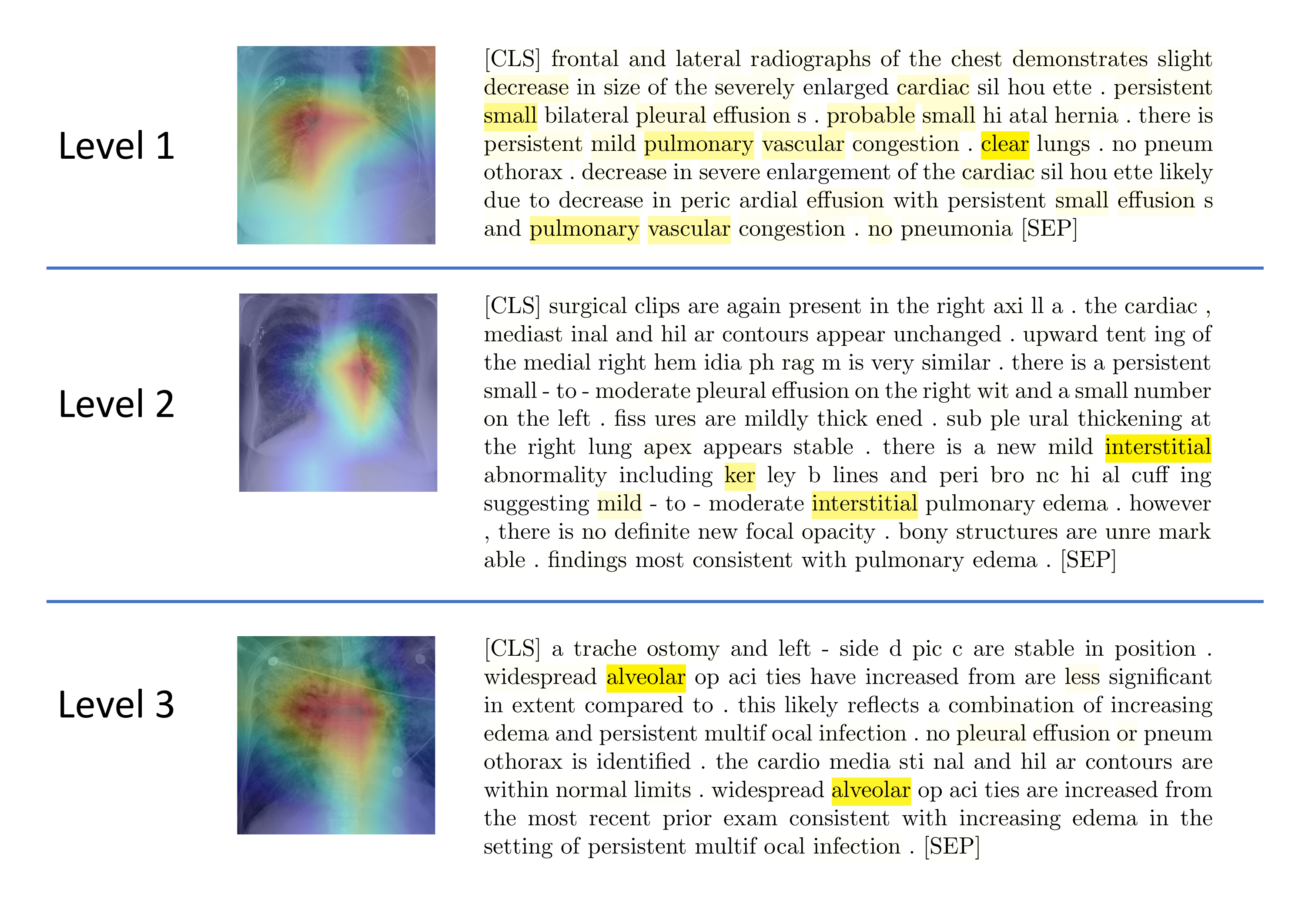}
	\caption{Joint model visualization. Top to bottom: (Level~1) The highlight of the Grad-CAM image is centered around the right hilar region, which is consistent with findings in pulmonary vascular congestion as shown in the report. (Level~2) The highlight of the Grad-CAM image is centered around the left hilar region which shows radiating interstitial markings as confirmed by the report heatmap. (Level 3) Grad-CAM highlights bilateral alveolar opacities radiating out from the hila and sparing the outer lungs.  This pattern is classically described as ``batwing" pulmonary edema mentioned in the report. The report text is presented in the form of sub-word tokenization performed by the BERT model, starting the report with a \texttt{[CLS]} token and ending with a \texttt{[SEP]}.}
	\label{fig:joint-viz}
\end{figure*}

\paragraph{\textbf{Joint Model Visualization.}}
As a by-product, our approach provides the possibility of interpreting model classification using text. While a method like Grad-CAM~\cite{selvaraju2017grad} can be used to localize regions in the image that are ``important" to the model prediction, it does not identify the relevant characteristics of the radiographs, such as texture. By leveraging the image-text embedding association, we visualize the heatmap of text attention corresponding to the last layer of the \texttt{[CLS]} token in the BERT model. This heatmap indicates report tokens that are important to our model prediction. As shown in Fig.~\ref{fig:joint-viz}, we use Grad-CAM~\cite{selvaraju2017grad} to localize relevant image regions and the highlighted words (radiographic findings, anatomical structures, etc.) from the text embedding to explain the model's decision making.

\section{Conclusion}

In this paper, we presented a neural network model that jointly learns from images and text to assess pulmonary edema severity from chest radiographs. The joint image-text representation learning framework incorporates the rich information present in the free-text radiology reports and significantly improves the performance of edema assessment compared to learning from images alone. Moreover, our experimental results show that joint representation learning benefits from the large amount of unlabeled image-text data.


Expert labeling of the radiology reports enabled us to quickly obtain a reasonable amount of test data, but this is inferior to direct labeling of images. The joint model visualization suggests the possibility of using the text to semantically explain the image model, which represents a promising direction for future investigation.

\subsubsection*{Acknowledgments.} This work was supported in part by NIH NIBIB NAC P41EB015902, Wistron Corporation, Takeda, MIT Lincoln Lab, and Philips. We also thank Dr. Daniel Moyer for helping generate Fig.~\ref{fig:joint-model}.



\bibliographystyle{splncs04}
\bibliography{paper1157}

\begin{thebibliography}{10}
\providecommand{\url}[1]{\texttt{#1}}
\providecommand{\urlprefix}{URL }
\providecommand{\doi}[1]{https://doi.org/#1}

\bibitem{adams2005characteristics}
Adams~Jr, K.F., Fonarow, G.C., Emerman, C.L., LeJemtel, T.H., Costanzo, M.R.,
  Abraham, W.T., Berkowitz, R.L., Galvao, M., Horton, D.P., Committee, A.S.A.,
  Investigators, et~al.: Characteristics and outcomes of patients hospitalized
  for heart failure in the united states: rationale, design, and preliminary
  observations from the first 100,000 cases in the acute decompensated heart
  failure national registry (adhere). American heart journal  \textbf{149}(2),
  209--216 (2005)

\bibitem{anderson2018bottom}
Anderson, P., He, X., Buehler, C., Teney, D., Johnson, M., Gould, S., Zhang,
  L.: Bottom-up and top-down attention for image captioning and visual question
  answering. In: Proceedings of the IEEE conference on computer vision and
  pattern recognition. pp. 6077--6086 (2018)

\bibitem{antol2015vqa}
Antol, S., Agrawal, A., Lu, J., Mitchell, M., Batra, D., Lawrence~Zitnick, C.,
  Parikh, D.: Vqa: Visual question answering. In: Proceedings of the IEEE
  international conference on computer vision. pp. 2425--2433 (2015)

\bibitem{beltagy2019scibert}
Beltagy, I., Lo, K., Cohan, A.: Scibert: A pretrained language model for
  scientific text. In: Proceedings of the 2019 Conference on Empirical Methods
  in Natural Language Processing and the 9th International Joint Conference on
  Natural Language Processing (EMNLP-IJCNLP). pp. 3606--3611 (2019)

\bibitem{ben2017mutan}
Ben-Younes, H., Cadene, R., Cord, M., Thome, N.: Mutan: Multimodal tucker
  fusion for visual question answering. In: Proceedings of the IEEE
  international conference on computer vision. pp. 2612--2620 (2017)

\bibitem{chapman2001simple}
Chapman, W.W., Bridewell, W., Hanbury, P., Cooper, G.F., Buchanan, B.G.: A
  simple algorithm for identifying negated findings and diseases in discharge
  summaries. Journal of biomedical informatics  \textbf{34}(5),  301--310
  (2001)

\bibitem{chechik2010large}
Chechik, G., Sharma, V., Shalit, U., Bengio, S.: Large scale online learning of
  image similarity through ranking. Journal of Machine Learning Research
  \textbf{11}(Mar),  1109--1135 (2010)

\bibitem{devlin2018bert}
Devlin, J., Chang, M.W., Lee, K., Toutanova, K.: Bert: Pre-training of deep
  bidirectional transformers for language understanding. arXiv preprint
  arXiv:1810.04805  (2018)

\bibitem{gheorghiade2010assessing}
Gheorghiade, M., Follath, F., Ponikowski, P., Barsuk, J.H., Blair, J.E.,
  Cleland, J.G., Dickstein, K., Drazner, M.H., Fonarow, G.C., Jaarsma, T.,
  et~al.: Assessing and grading congestion in acute heart failure: a scientific
  statement from the acute heart failure committee of the heart failure
  association of the european society of cardiology and endorsed by the
  european society of intensive care medicine. European journal of heart
  failure  \textbf{12}(5),  423--433 (2010)

\bibitem{goldberger2000physiobank}
Goldberger, A.L., Amaral, L.A., Glass, L., Hausdorff, J.M., Ivanov, P.C., Mark,
  R.G., Mietus, J.E., Moody, G.B., Peng, C.K., Stanley, H.E.: Physiobank,
  physiotoolkit, and physionet: components of a new research resource for
  complex physiologic signals. circulation  \textbf{101}(23),  e215--e220
  (2000)

\bibitem{harwath2018jointly}
Harwath, D., Recasens, A., Sur{\'\i}s, D., Chuang, G., Torralba, A., Glass, J.:
  Jointly discovering visual objects and spoken words from raw sensory input.
  In: Proceedings of the European Conference on Computer Vision (ECCV). pp.
  649--665 (2018)

\bibitem{harwath2016unsupervised}
Harwath, D., Torralba, A., Glass, J.: Unsupervised learning of spoken language
  with visual context. In: Advances in Neural Information Processing Systems.
  pp. 1858--1866 (2016)

\bibitem{he2016deep}
He, K., Zhang, X., Ren, S., Sun, J.: Deep residual learning for image
  recognition. In: Proceedings of the IEEE conference on computer vision and
  pattern recognition. pp. 770--778 (2016)

\bibitem{horng2020deep}
Horng, S., Liao, R., Wang, X., Dalal, S., Golland, P., Berkowitz, S.J.: Deep
  learning to quantify pulmonary edema in chest radiographs. arXiv preprint
  arXiv:2008.05975  (2020)

\bibitem{hunt20092009}
Hunt, S.A., Abraham, W.T., Chin, M.H., Feldman, A.M., Francis, G.S., Ganiats,
  T.G., Jessup, M., Konstam, M.A., Mancini, D.M., Michl, K., et~al.: 2009
  focused update incorporated into the acc/aha 2005 guidelines for the
  diagnosis and management of heart failure in adults: a report of the american
  college of cardiology foundation/american heart association task force on
  practice guidelines developed in collaboration with the international society
  for heart and lung transplantation. Journal of the American College of
  Cardiology  \textbf{53}(15),  e1--e90 (2009)

\bibitem{jing2017automatic}
Jing, B., Xie, P., Xing, E.: On the automatic generation of medical imaging
  reports. arXiv preprint arXiv:1711.08195  (2017)

\bibitem{johnson2019mimic}
Johnson, A.E., Pollard, T.J., Berkowitz, S.J., Greenbaum, N.R., Lungren, M.P.,
  Deng, C.y., Mark, R.G., Horng, S.: Mimic-cxr, a de-identified publicly
  available database of chest radiographs with free-text reports. Scientific
  Data  \textbf{6} (2019)

\bibitem{liao2019semi}
Liao, R., Rubin, J., Lam, G., Berkowitz, S., Dalal, S., Wells, W., Horng, S.,
  Golland, P.: Semi-supervised learning for quantification of pulmonary edema
  in chest x-ray images. arXiv preprint arXiv:1902.10785  (2019)

\bibitem{lu2016hierarchical}
Lu, J., Yang, J., Batra, D., Parikh, D.: Hierarchical question-image
  co-attention for visual question answering. In: Advances in neural
  information processing systems. pp. 289--297 (2016)

\bibitem{moradi2018bimodal}
Moradi, M., Madani, A., Gur, Y., Guo, Y., Syeda-Mahmood, T.: Bimodal network
  architectures for automatic generation of image annotation from text. In:
  International Conference on Medical Image Computing and Computer-Assisted
  Intervention. pp. 449--456. Springer (2018)

\bibitem{scispacy2019}
Neumann, M., King, D., Beltagy, I., Ammar, W.: {S}cispa{C}y: Fast and robust
  models for biomedical natural language processing. In: Proceedings of the
  18th BioNLP Workshop and Shared Task. pp. 319--327. Association for
  Computational Linguistics, Florence, Italy (Aug 2019).
  \doi{10.18653/v1/W19-5034}, \url{https://www.aclweb.org/anthology/W19-5034}

\bibitem{plummer2017enhancing}
Plummer, B.A., Brown, M., Lazebnik, S.: Enhancing video summarization via
  vision-language embedding. In: Proceedings of the IEEE conference on computer
  vision and pattern recognition. pp. 5781--5789 (2017)

\bibitem{selvaraju2017grad}
Selvaraju, R.R., Cogswell, M., Das, A., Vedantam, R., Parikh, D., Batra, D.:
  Grad-cam: Visual explanations from deep networks via gradient-based
  localization. In: Proceedings of the IEEE international conference on
  computer vision. pp. 618--626 (2017)

\bibitem{vasudevan2017query}
Vasudevan, A.B., Gygli, M., Volokitin, A., Van~Gool, L.: Query-adaptive video
  summarization via quality-aware relevance estimation. In: Proceedings of the
  25th ACM international conference on Multimedia. pp. 582--590 (2017)

\bibitem{attention2017allyouneed}
Vaswani, A., Shazeer, N., Parmar, N., Uszkoreit, J., Jones, L., Gomez, A.N.,
  Kaiser, L.u., Polosukhin, I.: Attention is all you need. In: Guyon, I.,
  Luxburg, U.V., Bengio, S., Wallach, H., Fergus, R., Vishwanathan, S.,
  Garnett, R. (eds.) Advances in Neural Information Processing Systems 30, pp.
  5998--6008. Curran Associates, Inc. (2017),
  \url{http://papers.nips.cc/paper/7181-attention-is-all-you-need.pdf}

\bibitem{wang2018tienet}
Wang, X., Peng, Y., Lu, L., Lu, Z., Summers, R.M.: Tienet: Text-image embedding
  network for common thorax disease classification and reporting in chest
  x-rays. In: Proceedings of the IEEE conference on computer vision and pattern
  recognition. pp. 9049--9058 (2018)

\bibitem{wolf2019huggingface}
Wolf, T., Debut, L., Sanh, V., Chaumond, J., Delangue, C., Moi, A., Cistac, P.,
  Rault, T., Louf, R., Funtowicz, M., et~al.: Huggingface's transformers:
  State-of-the-art natural language processing. ArXiv pp. arXiv--1910 (2019)

\bibitem{xu2015show}
Xu, K., Ba, J., Kiros, R., Cho, K., Courville, A., Salakhudinov, R., Zemel, R.,
  Bengio, Y.: Show, attend and tell: Neural image caption generation with
  visual attention. In: International conference on machine learning. pp.
  2048--2057 (2015)

\bibitem{xue2019improved}
Xue, Y., Huang, X.: Improved disease classification in chest x-rays with
  transferred features from report generation. In: International Conference on
  Information Processing in Medical Imaging. pp. 125--138. Springer (2019)

\end{thebibliography}


\begin{thebibliography}{8}
\bibitem{ref_article1}
Author, F.: Article title. Journal \textbf{2}(5), 99--110 (2016)

\bibitem{ref_lncs1}
Author, F., Author, S.: Title of a proceedings paper. In: Editor,
F., Editor, S. (eds.) CONFERENCE 2016, LNCS, vol. 9999, pp. 1--13.
Springer, Heidelberg (2016). \doi{10.10007/1234567890}

\bibitem{ref_book1}
Author, F., Author, S., Author, T.: Book title. 2nd edn. Publisher,
Location (1999)

\bibitem{ref_proc1}
Author, A.-B.: Contribution title. In: 9th International Proceedings
on Proceedings, pp. 1--2. Publisher, Location (2010)

\bibitem{ref_url1}
LNCS Homepage, \url{http://www.springer.com/lncs}. Last accessed 4
Oct 2017
\end{thebibliography}

\newpage

\section*{Supplementary Materials}

\begin{table}[!th]
    \centering
     \scalebox{0.95}{
  \begin{tabular}{|l|l|r|r|}
  \hline
  Edema severity & Regex keyword terms & Number of reports & Accuracy \\ \hline
  “Overall” & N/A & 485 & 89.69\% \\ \hline
  Level 0 -- & (no) pulmonary edema & 222 & 88.74\%  \\ \cline{2-4}
  none & (no) vascular congestion & 43 & 100.00\% \\ \cline{2-4}
  (n=216) & (no) fluid overload & 4 & 100.00\%  
  \\ \cline{2-4}
   & (no) acute cardiopulmonary process & 115 & 98.27\% \\ \hline
  Level 1 -- & cephalization & 17 & 94.12\%  \\ \cline{2-4}
  vascular congestion & pulmonary vascular congestion & 96 & 98.96\% \\ \cline{2-4}
  (n=98) & hilar engorgement & 3 & 100.00\%  
  \\ \cline{2-4}
   & vascular plethora & 13 & 100.00\%  
  \\ \cline{2-4}
   & pulmonary vascular prominence & 1 & 100.00\% \\ \cline{2-4}
   & pulmonary vascular engorgement & 8 & 87.50\% \\ \hline
  Level 2 -- & interstitial opacities & 30 & 73.33\%  \\ \cline{2-4}
  interstitial edema & kerley & 13 & 100.00\% \\ \cline{2-4}
  (n=105) & interstitial edema & 92 & 94.57\%  
  \\ \cline{2-4}
   & interstitial thickening & 6 & 66.67\%
  \\ \cline{2-4}
   & interstitial pulmonary edema & 21 & 100.00\% \\ \cline{2-4}
   & interstitial marking & 19 & 68.42\% \\ \cline{2-4}
   & interstitial abnormality & 10 & 70.00\% \\ \cline{2-4}
   & interstitial abnormalities & 2 & 100.00\% \\ \cline{2-4}
   & interstitial process & 2 &
   100.00\% \\ \hline
  Level 3 -- & alveolar infiltrates & 10 & 100.00\%  \\ \cline{2-4}
  alveolar edema & severe pulmonary edema & 58 & 98.28\% \\ \cline{2-4}
  (n=66) & perihilar infiltrates & 1 & 100.00\% \\ \cline{2-4}
   & hilar infiltrates & 1 & 100.00\% \\ \cline{2-4}
   & parenchymal opacities & 6 & 16.67\% \\ \cline{2-4}
   & alveolar opacities & 7 & 100.00\% \\ \cline{2-4}
   & ill defined opacities & 1 & 100.00\% \\ \cline{2-4}
   & ill-defined opacities & 1 & 0.00\% \\ \cline{2-4}
   & patchy opacities & 10 & 10.00\% \\ \hline
  \end{tabular}
    }
    \vspace{0.2in}
    \caption*{Supplemental Table 1: Validation of regex keyword terms. The accuracy (positive predictive value) of the regular expression results for levels 0-3 based on the expert review results are 90.74\%, 80.61\%, 95.24\%, and 90.91\%, respectively. The total number of reports from all the keywords is more than 485 because some reports contain more than one keyword.}
    \label{tab:regex-words}
\end{table}

\newpage

\begin{table}[!h]
    \centering
     \scalebox{1}{
    \begin{tabular}{l r}
    \hline \hline
         \textbf{Hyperparameter} & \textbf{Setting}  \\ \hline
         number-of-epochs (supervised) & 50, 100, 150, \textbf{250} \\
         learning-rate & \textbf{2e-5}, 5e-4, 1e-4, 1e-3 \\
         learning-rate-scheduler & \small \textbf{warmup-linear}, reduce-on-plateau \\ \hline \hline
    \end{tabular}
    }
    \vspace{0.2in}
    \caption*{Supplemental Table 2: Hyper-parameter search. Hyper-parameter settings were firstly experimented on the joint model in a supervised learning fashion. The experiments were performed on 5-fold cross validation within the training set, while holding out the test set. A learning rate of 2e-5 and the warmup-linear scheduler were chosen. Finally, the number of epochs was further experimented for the semi-supervised joint model learning with the 5-fold cross validation.}
    \label{tab:hyperparam-search}
\end{table}

\begin{figure}[!ht]
	\centering
	\includegraphics[width=1\linewidth]{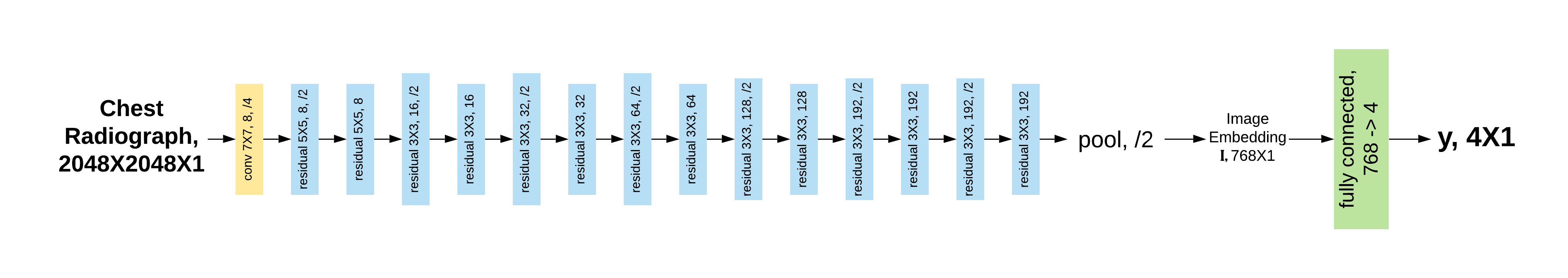}
	\includegraphics[width=0.7\linewidth]{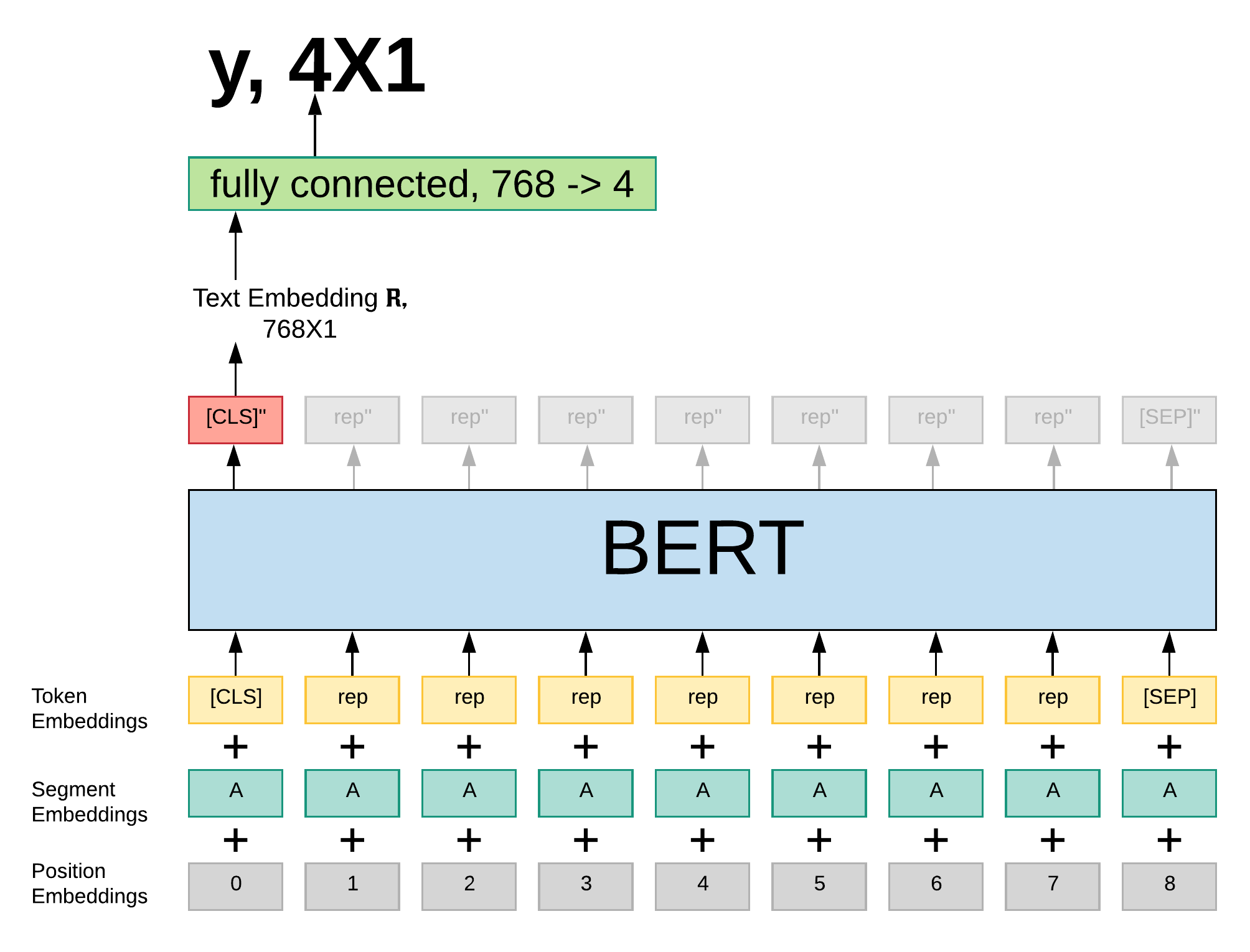}
         \vspace{0.2in}
	\caption*{Supplemental Figure 1: Top: Image encoder and classifier architecture. Each residual block includes 2 convolutional layers. Bottom: Text encoder and classifier architecture using the BERT model. A full radiology report is encoded between \texttt{[CLS]} and \texttt{[SEP]} tokens; \texttt{rep} is the text associated with the report. Maximum input sequence length is set to 320.}
	\label{fig:architecture}
\end{figure}

\newpage

\begin{figure}[!hb]
\begin{minipage}[c]{\linewidth}
    \centering
	\includegraphics[width=0.49\linewidth]{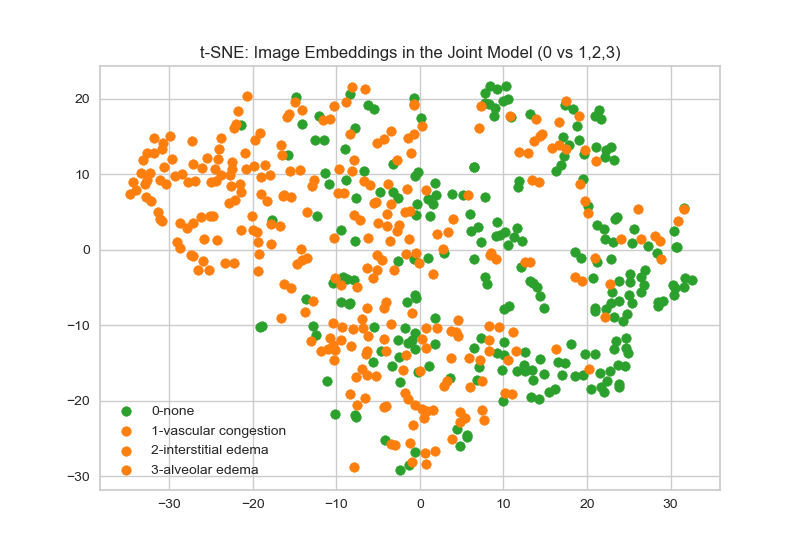}
	\includegraphics[width=0.49\linewidth]{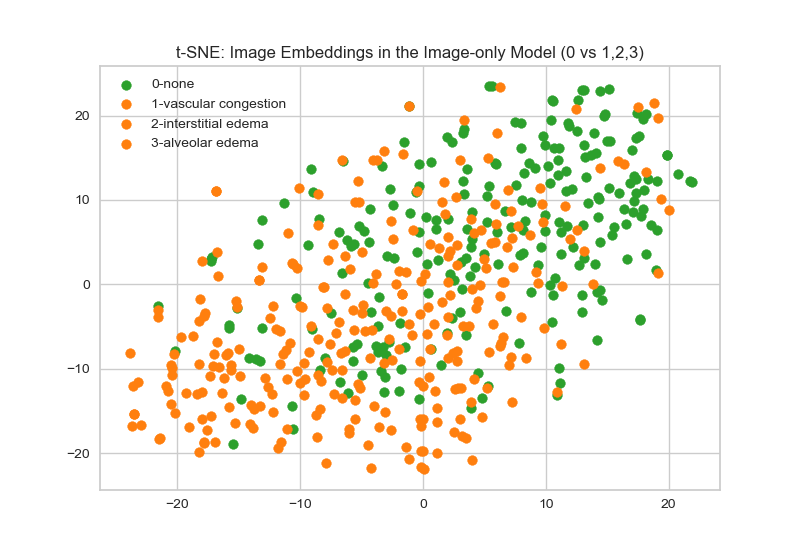}
	
	\includegraphics[width=0.49\linewidth]{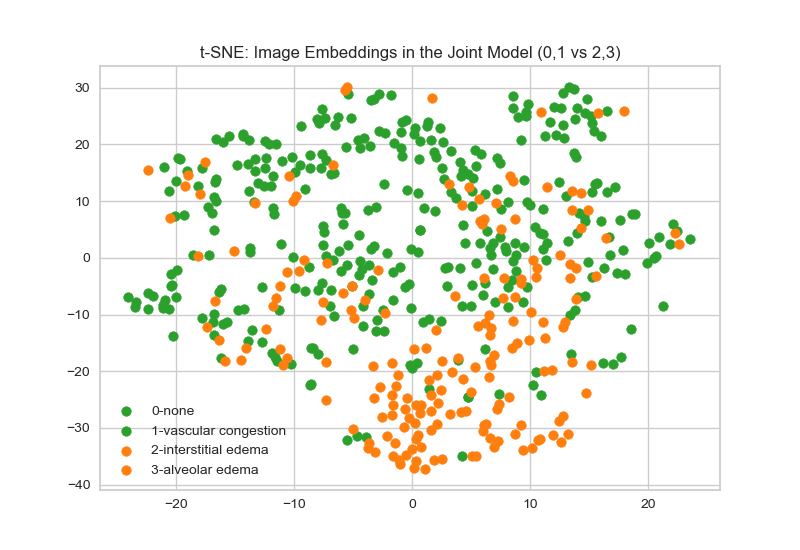}
	\includegraphics[width=0.49\linewidth]{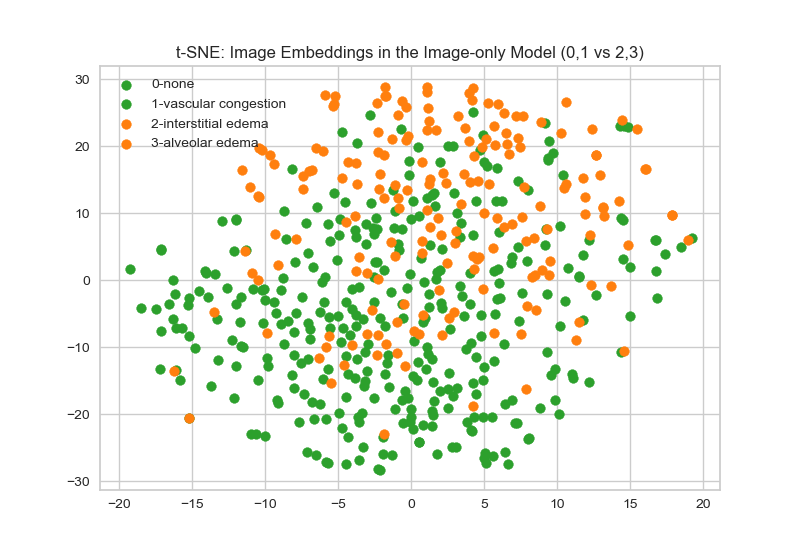}
	
	\includegraphics[width=0.49\linewidth]{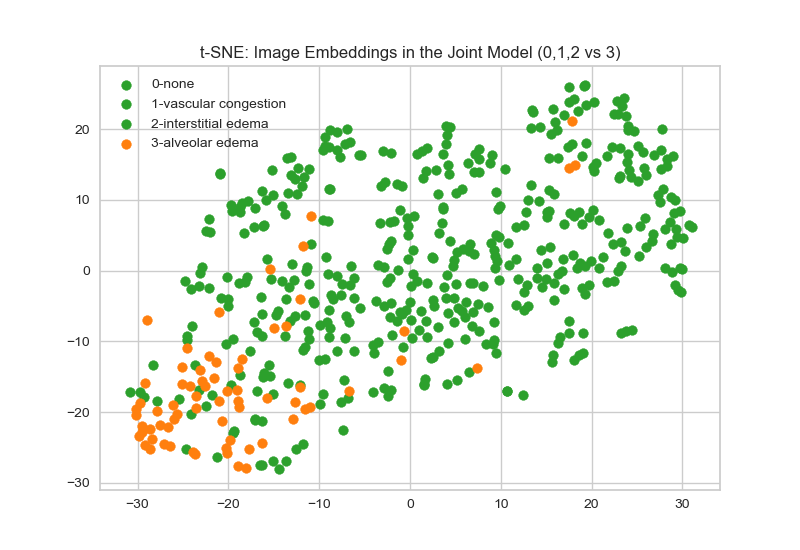}
	\includegraphics[width=0.49\linewidth]{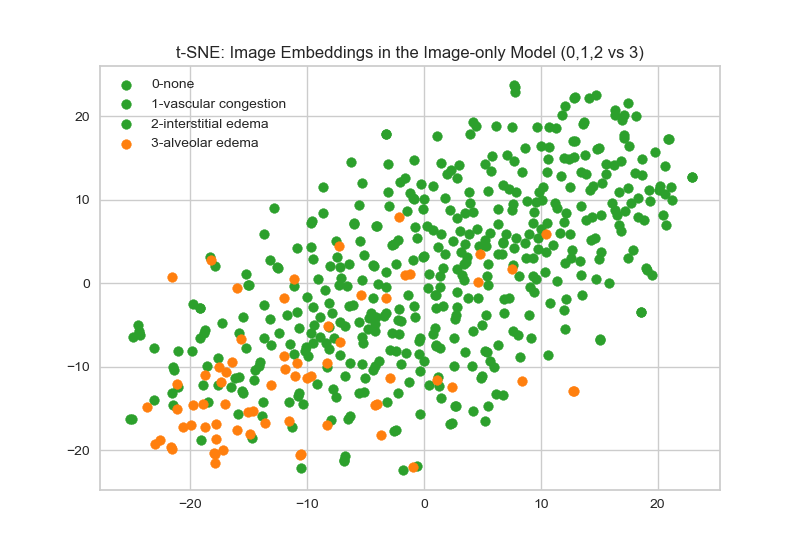}
	\label{fig:image-model}
\end{minipage}
 \vspace{0.4in}
\begin{minipage}[c]{\linewidth}
	\raggedright
	Supplemental Figure 2: t-SNE visualization in 2 dimensions for image embeddings in the joint model (left) and the embeddings in the image-only model (right).
\end{minipage}
\end{figure}

\end{document}